\documentclass[10pt,twocolumn,letterpaper]{article}

\usepackage{cvpr}
\usepackage{times}
\usepackage{epsfig}
\usepackage{graphicx}
\usepackage{amsmath}
\usepackage{amssymb}

\usepackage{xcolor}
\usepackage{cite}
\usepackage{color}
\usepackage{subfigure}
\usepackage{booktabs}
\usepackage{bm}
\usepackage{stfloats}% for figure visualize
\usepackage[title]{appendix}% for appendix
\usepackage{pifont}% http://ctan.org/pkg/pifont
\newcommand{\cmark}{\ding{51}}%
\newcommand{\xmark}{\ding{55}}%

\usepackage[pagebackref=true,breaklinks=true,letterpaper=true,colorlinks,bookmarks=false]{hyperref}

\cvprfinalcopy % *** Uncomment this line for the final submission

\def\cvprPaperID{3547} % *** Enter the CVPR Paper ID here

\ifcvprfinal\fi
\begin{document}

%%%%%%%%% TITLE
\title{Attention-guided Unified Network for Panoptic Segmentation}

\author{Yanwei Li$^{1,2}$, Xinze Chen$^{3}$, Zheng Zhu$^{1,2}$, Lingxi Xie$^{4,5}$, Guan Huang$^{3}$, \\
Dalong Du$^{3}$, Xingang Wang$^{1}$  \\
$^{1}$Institute of Automation, CAS\quad$^{2}$University of Chinese Academy of Sciences\\
$^{3}$Horizon Robotics, Inc.\quad$^{4}$Johns Hopkins University\quad$^{5}$Noah's Ark Lab, Huawei Inc.\\
{\tt\small \{liyanwei2017,zhuzheng2014,xingang.wang\}@ia.ac.cn}\\
{\tt\small \{xinze.chen,guan.huang,dalong.du\}@horizon.ai}\quad{\tt\small 198808xc@gmail.com}
}

\maketitle
\thispagestyle{empty}

%%%%%%%%% ABSTRACT
\begin{abstract}
This paper studies panoptic segmentation, a recently proposed task which segments foreground (FG) objects at the instance level as well as background (BG) contents at the semantic level. Existing methods mostly dealt with these two problems separately, but in this paper, we reveal the underlying relationship between them, in particular, FG objects provide complementary cues to assist BG understanding. Our approach, named the Attention-guided Unified Network (AUNet), is a unified framework with two branches for FG and BG segmentation simultaneously. Two sources of attentions are added to the BG branch, namely, RPN and FG segmentation mask to provide object-level and pixel-level attentions, respectively. Our approach is generalized to different backbones with consistent accuracy gain in both FG and BG segmentation, and also sets new state-of-the-arts both in the MS-COCO (46.5\% PQ) and Cityscapes (59.0\% PQ)  benchmarks.
\end{abstract}

%%%%%%%%% BODY TEXT
\section{Introduction}
Scene understanding is a fundamental yet challenging task in computer vision, which has a great impact on other applications such as autonomous driving and robotics. Classic tasks for scene understanding mainly include object detection, instance segmentation and semantic segmentation. This paper considers a recently proposed task named {\em panoptic segmentation}~\cite{kirillov2018panoptic}, which aims at finding all foreground (FG) objects (named {\em things}, mainly including countable targets such as {\em people}, {\em animals}, {\em tools}, {\em etc.}) at the instance level, meanwhile parsing the background (BG) contents (named {\em stuff}, mainly including amorphous regions of similar texture and/or material such as {\em grass}, {\em sky}, {\em road}, {\em etc.}) at the semantic level. The benchmark algorithm~\cite{kirillov2018panoptic} and MS-COCO panoptic challenge winners~\cite{coco2018lead} dealt with this task by directly combining FG instance segmentation models~\cite{he2017mask} and BG scene parsing~\cite{zhao2017pyramid} algorithms, which ignores the underlying relationship and fails to borrow rich contextual cues between {\em things} and {\em stuff}.
%------------------------------------------------------------------------
\begin{figure}[t!] \label{fig_intro}
\centering
\subfigure[Input Image]{
\includegraphics[width=0.43\linewidth]{Fig/intro_fig_raw.jpg}
\label{fig_intro_raw}
}
\quad
\subfigure[Panoptic Segmentation]{
\includegraphics[width=0.43\linewidth]{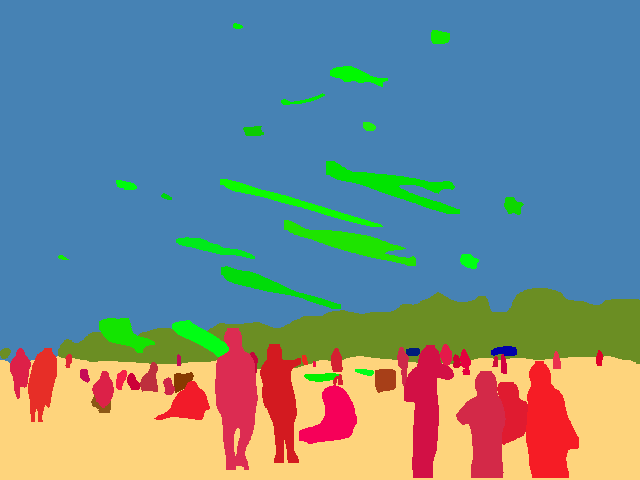}
\label{fig_intro_all}
}
\quad
\subfigure[Foreground: {\em things}]{
\includegraphics[width=0.43\linewidth]{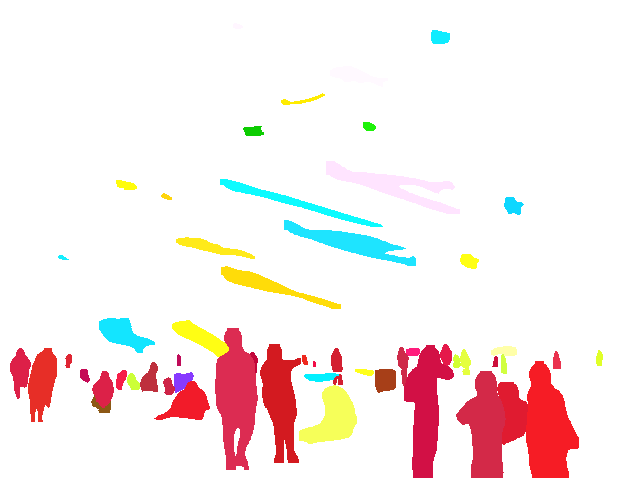}
\label{fig_intro_thing}
}
\quad
\subfigure[Background: {\em stuff}]{
\includegraphics[width=0.43\linewidth]{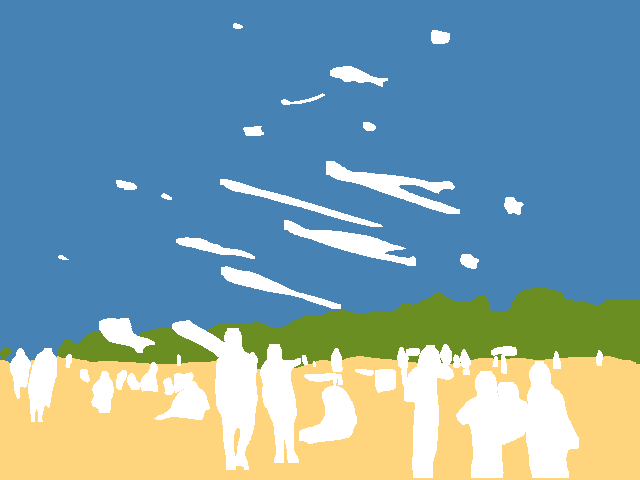}
\label{fig_intro_stuff}
}
\caption{Given an image~\ref{fig_intro_raw}, the goal of panoptic segmentation~\ref{fig_intro_all} is to find FG {\em things} at the instance level~\ref{fig_intro_thing} and BG {\em stuff} at the semantic level~\ref{fig_intro_stuff}. The {\em things} of the same class share the same color family but appear in different intensities.  {\bf All these results are produced by the proposed approach.}}
\end{figure}
%------------------------------------------------------------------------

In this paper, we present a conceptually simple and unified framework for panoptic segmentation. To facilitate information flow between FG {\em things} and BG {\em stuff}, we combine conventional instance segmentation and semantic segmentation networks, leading to a unified network with two branches. This strategy brings an immediate improvement in segmentation accuracy as well as higher efficiency in computation (because the network backbone can be shared). This implies that panoptic segmentation benefits from complementary information provided by FG objects and BG contents, which lays the foundation of our approach.

Going one step further, we explore the possibility of integrating higher-level visual cues ({\em i.e.}, beyond the features extracted from the end of the backbone) towards the more accurate segmentation. This is achieved via two attention-based modules working at the object level and the pixel level, respectively. For the first module, we refer to the regional proposals, each of which indicates a possible FG {\em thing}, and adjusts the probability of the corresponding region to be considered as FG {\em things} and BG {\em stuff}. For the second module, we take out the FG segmentation mask, and use it to refine the boundary between FG {\em things} and BG {\em stuff}. In the context of deep networks, these two modules, named the Proposal Attention Module (PAM) and Mask Attention Module (MAM), respectively, are implemented as additional connections across FG and BG branches. Within MAM, a new layer named {\em RoIUpsample} is designed to define an accurate mapping function between pixels in the fixed-shape FG mask and the corresponding feature map. In practice, all additional connections go from the FG branch to the BG branch, mainly due to the observation that FG segmentation is often more accurate\footnote{We find the {\em pixel accuracy} of {\em things} is much higher (6.7\% absolute gap) than that of {\em stuff}, when considering instance with the same semantic as one category, {\em e.g.}, all individuals are evaluated as {\em person} in testing. We evaluate them on the same MS-COCO semantic evaluation metric.\label{footnote_pri_info}}. Furthermore, BG {\em stuff}, while being refined by FG {\em things}, also gives feedback via gradients. Consequently, both FG and BG segmentation accuracies are considerably improved.

The overall approach, named Attention-guided Unified Network ({\bf AUNet}), can be easily instantiated to various network backbones, and optimized in an end-to-end manner. We evaluate AUNet in two popular segmentation benchmarks, namely, the MS-COCO~\cite{lin2014microsoft} and Cityscapes~\cite{cordts2016cityscapes} datasets, and claim {\bf the state-of-the-art performance} in terms of PQ, a standard metric integrating accuracies of both {\em things} and {\em stuff}~\cite{kirillov2018panoptic}. In addition, the benefits brought by joint optimization and two attention-based modules are verified through an extensive ablation study~\ref{sec::abla_study}.

The major contribution of this research is to present a simple and unified framework for both FG and BG segmentation, which reaches the top performance in MS-COCO~\cite{lin2014microsoft} and Cityscapes~\cite{cordts2016cityscapes} datasets. Furthermore, this work also investigate the complementary information delivered by FG objects and BG contents. While panoptic segmentation serves as a natural scenario of studying this topic, its application lies in a wider range of visual tasks. Our solution, AUNet, is a preliminary exploration in this field, yet we look forward to more efforts along this direction.

The remainder of this paper is organized as follows. Section~\ref{sec:related_work} briefly reviews related work. Section~\ref{sec:approach} elaborates the proposed AUNet, including two attention-based modules. After experiments are shown in Section~\ref{sec:experiments}, we conclude this work in Section~\ref{sec:conclusions}.

%------------------------------------------------------------------------
\section{Related Work}
\label{sec:related_work}
Traditional deep learning based scene understanding researches often focused on foreground or background targets~\cite{he2017mask, zhao2017pyramid}. Recently, the rapid progress in object detection~\cite{girshick2014rich, girshick2015fast, ren2015faster} and instance segmentation~\cite{dai2016instance, li2016fully, he2017mask, liu2018path} made it possible to achieve object localization and segmentation at a finer level. Meanwhile, the development of semantic segmentation~\cite{long2015fully, zhao2017pyramid, chen2018deeplab, chen2017rethinking} boosted the performance of scene parsing. Despite their effectiveness, the separation of these tasks caused the lack of contextual cues in instance segmentation as well as the confusion brought by individuals in semantic segmentation. To bridge this gap, recently, researchers proposed a new task named {\em panoptic segmentation}~\cite{kirillov2018panoptic}, which aims at accomplishing both tasks (FG instance and BG semantic segmentation) simultaneously.
%-------------------------------------------------------------------------

\vspace{0.05cm}
\noindent{\bf Panoptic Segmentation:} In \cite{kirillov2018panoptic}, the author gave a benchmark of panopic segmentation by combining instance and semantic segmentation models. Later, a weakly-supervised method~\cite{li2018weakly} was proposed on top of initialized semantic results, and an end-to-end approach~\cite{de2018panoptic} was designed to combine both FG and BG cues. However, their performance is far from the benchmark \cite{kirillov2018panoptic}. Different from them, our proposed AUNet achieves the top performance in an end-to-end framework. Furthermore, we also establish the bond between proposal-based instance and FCN based semantic segmentation. Most recently works include~\cite{kirillov2019panoptic, xiong2019upsnet, liu2019end}.
%-------------------------------------------------------------------------

\vspace{0.05cm}
\noindent{\bf Instance Segmentation:} Instance segmentation aims at discriminating different instances of the same object. There are mainly two streams of methods to solve this task, namely, proposal-based methods and segmentation-based methods. Proposal-based methods, with the help of accurate regional proposals, often achieved higher performance. Recent examples include MNC~\cite{dai2016instance}, FCIS~\cite{li2016fully}, Mask R-CNN~\cite{he2017mask} and~PANet \cite{liu2018path}. Moreover, segmentation-based methods aggregated pixel-level cues to compose instances combined with semantic segmentation~\cite{liang2015proposal, arnab2017pixelwise, liu2018affinity} or depth ordering~\cite{zhang2016instance} results.
%-------------------------------------------------------------------------

\vspace{0.05cm}
\noindent{\bf Semantic Segmentation:} With the development of so-called encoding-decoding networks such as FCN~\cite{long2015fully}, rapid progress has been made in semantic segmentation~\cite{zhao2017pyramid, chen2018deeplab, chen2017rethinking}. In segmentation, capturing contextual information plays a vital role, for which various approaches were proposed including ASPP used in DeepLab~\cite{chen2018deeplab, chen2017rethinking} for multi-scale contexts, DenseASPP~\cite{yang2018denseaspp} for global contexts, and PSPNet~\cite{zhao2017pyramid} which collected contextual priors. There were also efforts to use attention modules for spatial feature selection, such as~\cite{yu2018learning, zhang2018context, fu2018dual}, which will be detailed discussed next.
%-------------------------------------------------------------------------

\begin{figure*}[th!]
  \centering
  \includegraphics[width=0.99\linewidth]{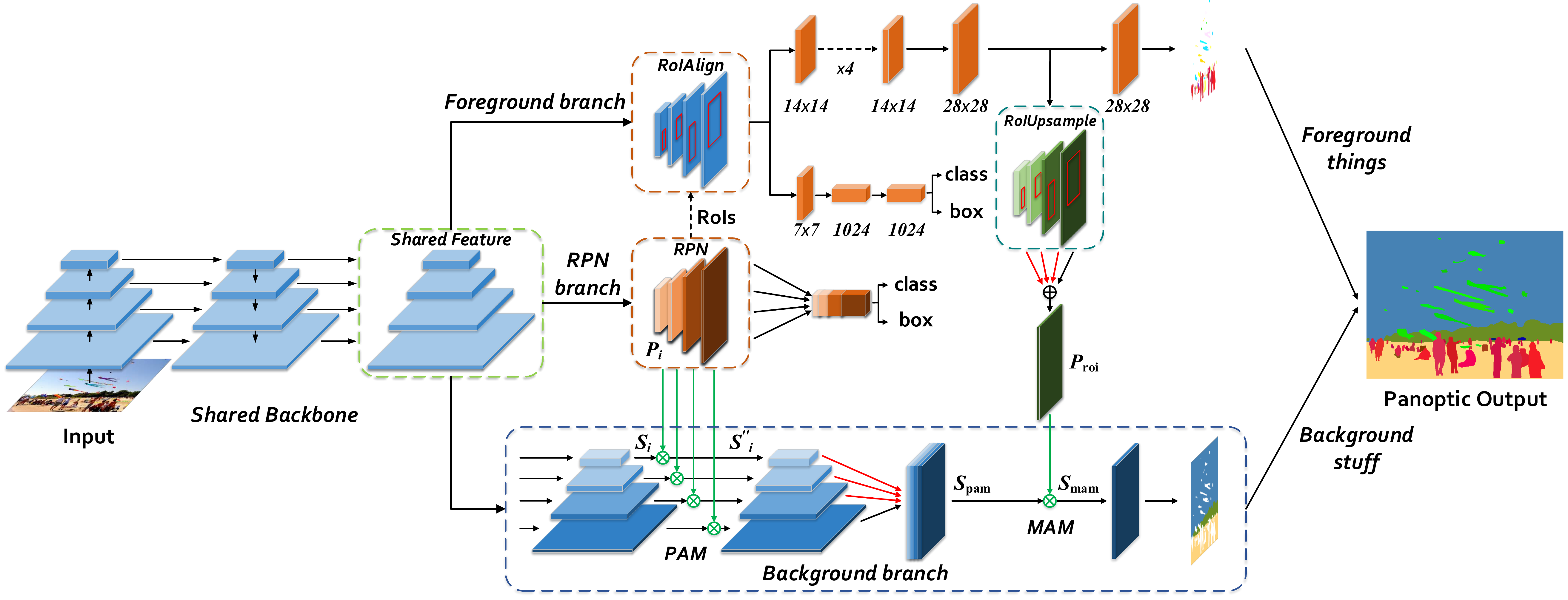}\\
  \caption{The proposed network structure. We adopt FPN as our backbone and share features with three parallel branches, namely {\em foreground branch}, {\em background branch}, and {\em RPN branch}. In the training stage, the network is optimized in an end-to-end manner. In the inference stage, panoptic results are generated by {\em things} and {\em stuff} results following the method described in Section \ref{sec::implem_detail}. ``$\oplus$'' denotes element-wise sum and  the green ``$\otimes$'' represents Proposal Attention Module (PAM) or Mask Attention Module (MAM) according to its position. PAM and MAM model the complementary relation between two branches. Details of PAM and MAM are shown in Figure \ref{fig_rpn_att} and Figure \ref{fig_mask_att}. The red and green arrows represent upsample and attention operations, respectively.}\label{fig_arch}
\end{figure*}
%-------------------------------------------------------------------------

\vspace{0.05cm}
\noindent{\bf Attention-based Modules:} Attention-based modules have been widely applied in visual tasks, including image processing, video understanding, and object tracking~\cite{hu2017squeeze, chen2016attention, wang2018non, Zhu_2018_ECCV, Zhu_2018_CVPR}. In particular, SENet~\cite{hu2017squeeze} formulated channel-wise relationships via an attention-and-gating mechanism, non-local network~\cite{wang2018non} bridged self-attention for machine translation~\cite{vaswani2017attention} to video classification using non-local filters. In the scope of scene understanding, \cite{yu2018learning} and~\cite{zhang2018context} aggregated global contextual information as well as class-dependent features by channel-attention operations. More recently, self-attention and channel attention were adopted by~\cite{fu2018dual} to model long-range contexts in the spatial and channel dimensions, respectively. In this work, we establish the relationship between foreground {\em things} and background {\em stuff} in panoptic segmentation with a series of coarse-to-fine attention blocks.
%-------------------------------------------------------------------------

\section{Attention-guided Unified Network}
\label{sec:approach}

\subsection{Problem and Baselines}
Panoptic segmentation task aims at understanding everything visible in one view, which means each pixel of an image must be assigned a semantic label and an instance ID. To address this issue, the existing top algorithms~\cite{kirillov2018panoptic, coco2018lead} directly combined the instance and semantic results from separate models, such as Mask R-CNN~\cite{he2017mask} and PSPNet~\cite{zhao2017pyramid}.

We formulate the problem of panoptic segmentation as recognizing and segmenting all FG {\em things} and understanding all BG {\em stuff}. In this way, we solve the problem from two aspects, namely foreground branch and background branch in a unified network (Figure \ref{fig_arch}). In detail, given an input image $\bm X$, our goal is to generate FG {\em things} result $\bm Y_{{\rm{Th}}}$ and BG {\em stuff} result $\bm Y_{{\rm{St}}}$ simultaneously. Thus, the panoptic result $\bm Y_{{\rm{Pa}}}$ can be generated from $\bm Y_{{\rm{Th}}}$ and $\bm Y_{{\rm{St}}}$ directly using the fusion method in Section~\ref{sec::implem_detail}. The performance of panoptic results is evaluated by panoptic quality (PQ) \cite{kirillov2018panoptic} as described in Section \ref{eval_metric}. For this purpose, we firstly introduce our unified framework for panoptic segmentation in this section. Then, key elements in our designed attention-guided modules are elaborated, including proposal attention module (PAM) and mask attention module (MAM). Finally, we give our implementation details.

In this work, we view the method, in which {\em things} and {\em stuff} are generated from separate models, as our baseline. Specifically, the baseline method gives the result of {\em things} $\bm Y_{{\rm{Th}}}$ and {\em stuff} $\bm Y_{{\rm{St}}}$ from separate models $\mathbb{M}_{{\rm{Th}}}$ and $\mathbb{M}_{{\rm{St}}}$ respectively. And the FG model $\mathbb{M}_{{\rm{Th}}}$ and BG model $\mathbb{M}_{{\rm{St}}}$ are given the similar backbones ({\em e.g.}, FPN~\cite{lin2017feature}) for the following unified framework.
%-------------------------------------------------------------------------
\subsection{Unified Framework} \label{sec::net_struct}
In order to bridge the gap between FG {\em things} with BG {\em stuff}, we propose the {\em Attention-guided Unified Network} (AUNet). Comparing with the baseline approach, the proposed AUNet fuses two models ($\mathbb{M}_{{\rm{Th}}}$ and $\mathbb{M}_{{\rm{St}}}$) together by sharing the same backbone and generates $\bm Y_{{\rm{Th}}}$ and $\bm Y_{{\rm{St}}}$ from parallel branches. As clearly illustrated in Figure~\ref{fig_arch}, the AUNet is conceptually simple: FPN is adopted as the backbone to extract discriminative features from different scales and shared by all the branches.

Different from traditional approaches, which directly combine results from $\mathbb{M}_{{\rm{Th}}}$ and $\mathbb{M}_{{\rm{St}}}$, the proposed AUNet optimizes them using a joint loss function $\mathcal{L}$ (defined in Section~\ref{sec::implem_detail}) and facilitates both tasks in a unified framework. In detail, we adopt a proposal-based instance segmentation module to generate finer masks $M$ in {\em foreground branch}. And for {\em background branch}, light heads are designed to aggregate scene information from shared multi-scale features. In this way, the shared backbone is supervised by FG {\em things} and BG {\em stuff} simultaneously, which promotes the connection between two branches in feature space. In order to build up the bond between FG objects and BG contents more explicitly, two sources of attention modules are added. We consider the coarse attention operation between the $i$-th scale BG feature map with the corresponding RPN feature map, denoted by $S_i$ and $P_i$ respectively. The attention module can be formulated as ${S_i} \otimes {P_i}$ , where ``$\otimes$'' denotes attention operations, as illustrated in Figure~\ref{fig_arch}. Furthermore, the finer relationship is established by the attention between the processed feature map $S_{{\rm{pam}}}$ and the generated FG segmentation mask $P_{{\rm{roi}}}$, which can be formulated as $S_{{\rm{pam}}} \otimes P_{{\rm{roi}}}$. Details will be investigated in the following section.
%-------------------------------------------------------------------------
\subsection{Attention-guided Modules} \label{sec::att_module}
Considering the complementary relationship between FG {\em things} and BG {\em stuff}, we introduce features from {\em foreground branch} to {\em background branch} for more contextual cues. From another perspective, the attention operation connecting two branches also establishes a bond between proposal-based method and FCN-based method segmentation. To this end, two spatial attention modules are proposed, namely proposal attention module (PAM) and mask attention module (MAM).
%------------------------------------------------------------------------
\subsubsection{Proposal Attention Module} \label{sec::rpn_att_sec}
%------------------------------------------------------------------------
\begin{figure}[b!]
  \centering
  \includegraphics[width=0.99\linewidth]{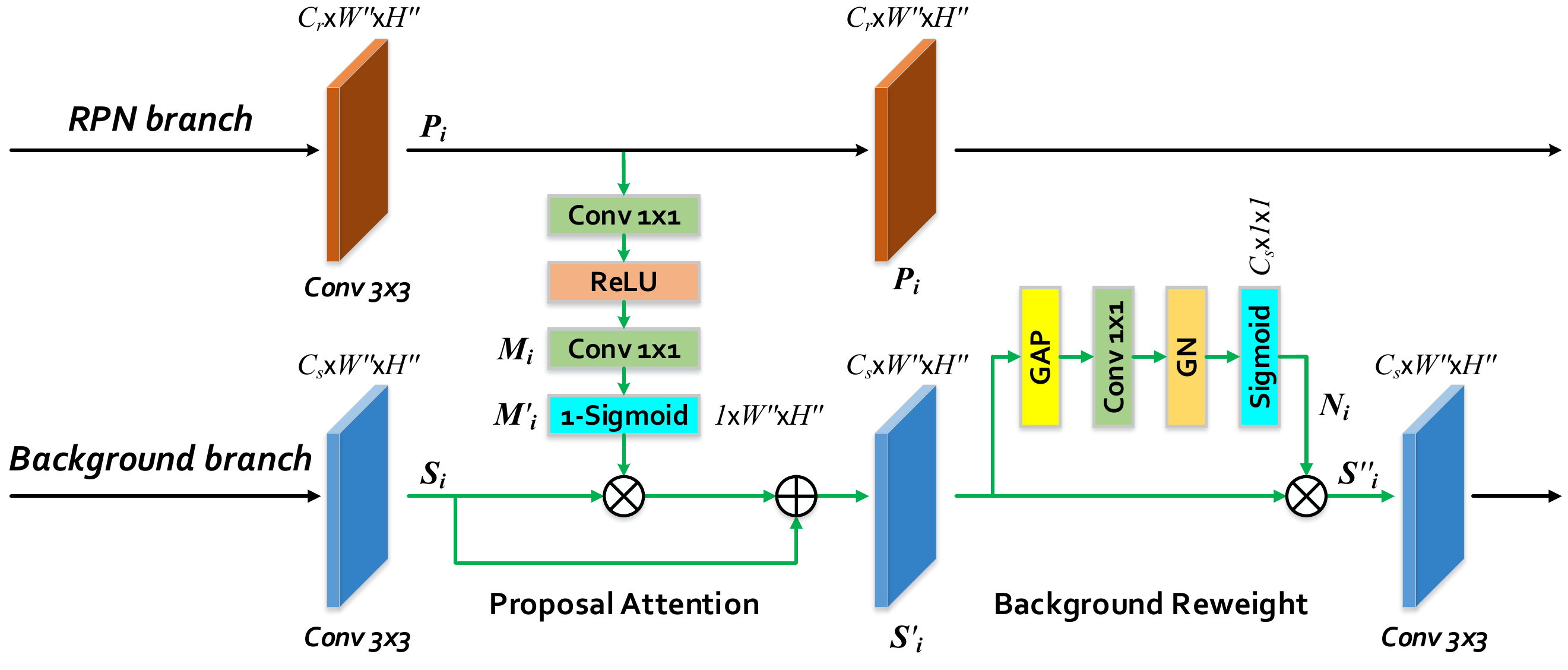}\\
  \caption{The designed proposal attention module (PAM) for complementary relationship establishment. We adopt this block in each scale of shared features, {\em i.e.}, $W''$ and $H''$ changes in each scale. Here, ``$\otimes$'' denotes spatial element-wise multiplication and ``$\oplus$'' denotes element-wise sum. The green arrows represent operations in PAM. GAP and GN indicate Global Average Pooling and Group Normalization \cite{Wu_2018_ECCV}, respectively.}\label{fig_rpn_att}
\end{figure}
%-------------------------------------------------------------------------
In classic two-stage detection frameworks, region proposal network (RPN) \cite{ren2015faster} is introduced to give predicted binary class labels (foreground and background) and bounding-box coordinates. This means RPN features contain rich background information which can only be obtained from {\em stuff} annotations in background branch. Therefore, we propose a new approach to establish the complementary relationship between FG elements and BG contents, called Proposal Attention Module (PAM). As shown in Figure \ref{fig_rpn_att}, we utilize contextual cues from RPN branch for attention operation. Here, we give a detailed formulation for this process. Given an input feature map $P_i \in {\mathbb{R}^{C_r \times W'' \times H''}}$ from the $i$-th scale RPN branch, the FG weighted map $M_i$ before $\mathrm{sigmoid}$ activation can be formulated as:
%-------------------------------------------------------------------------
\begin{equation} \label{equ_rpn_generate_mask}
{M_i} = f(\sigma (f({P_i},{w_{i,{\rm{1}}}})),{w_{i,{\rm{2}}}})
\end{equation}
%-------------------------------------------------------------------------
where  $f( \cdot , \cdot )$ denotes a convolution function, $\sigma$ represents the ReLU activation function, ${M_i} \in {\mathbb{R}^{1 \times W'' \times H''}}$ means the generated FG weighted map, both ${w_{i,1}} \in {\mathbb{R}^{C'_r \times C_r \times 1 \times 1}}$ and ${w_{i,2}} \in {\mathbb{R}^{1 \times C'_r \times 1 \times 1}}$ indicate convolutional parameters. 

To emphasize the background contents, we formulate the attention weighted map $M'_i$ as ${1 - \mathrm{sigmoid}({M_i})}$. Then, the $i$-th scale activated feature map $S'_{i} \in {\mathbb{R}^{ C_s \times W'' \times H''}}$ can be presented as:
\begin{equation}
S'_{i,j} = {S_{i,j}} \otimes M'_i \oplus {S_{i,j}}
\end{equation}
where $\otimes$ and $\oplus$ denotes element-wise multiplication and sum respectively, $S_{i,j}$ means the $j$-th layer of semantic feature map ${S_i} \in {\mathbb{R}^{ C_s \times W'' \times H''}}$. 

Motivated by~\cite{hu2017squeeze}, a simple background reweight function is designed to downweight useless background layers after attention operation. We believe it could be improved, but it is beyond the scope of this work. The reweighted feature map $S''_i \in {\mathbb{R}^{ C_s \times W'' \times H''}}$ can be generated as:
\begin{equation}\label{equ::pam_gap}
{N_i} = \mathrm{sigmoid}({\mathrm {GN}}(f({\mathrm G}(S{'_i}),{w_{i,3}})))
\end{equation}
\begin{equation}
S'{'_{i,k}} = S{'_{i,k}} \otimes {N_i}
\end{equation}
%-------------------------------------------------------------------------
where $\mathrm G$ and $\mathrm {GN}$ denotes global average pooling and group norm \cite{Wu_2018_ECCV} respectively, ${N_i} \in {\mathbb{R}^{{C_s} \times 1 \times 1}}$ means reweighting operator, ${w_{i,3}} \in {\mathbb{R}^{{C_s} \times {C_s} \times 1 \times 1}}$ represents convolutional parameter, and $S{'_{i,k}}$ indicates the $k$-th pixel channel in $S{'_{i}}$.

Based on the above formulation of PAM, we highlight the background regions in the shared feature maps via attention operation and background reweight function. It also facilitates the learning of {\em things} in turn by enhancing the weights of activated foreground regions during backpropagation (see Section \ref{sec::abla_study}).

%------------------------------------------------------------------------
\subsubsection{Mask Attention Module} \label{sec::mask_att_sec}
\begin{figure}[tb!]
  \centering
  \includegraphics[width=0.99\linewidth]{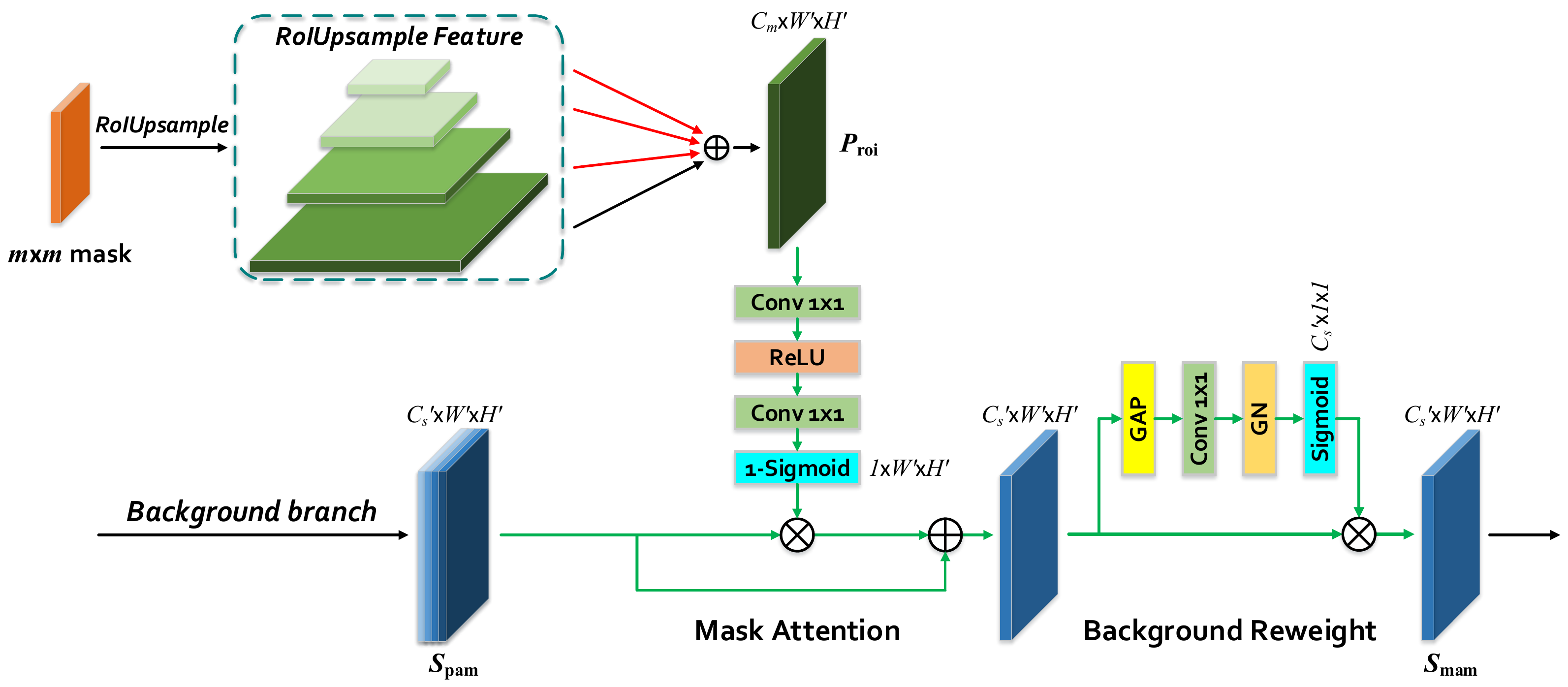}\\
  \caption{The proposed mask attention module (MAM) for a finer relationship modelling. Here, ``$\otimes$'' denotes spatial element-wise multiplication and ``$\oplus$'' denotes element-wise sum. The red and green arrows represent upsample and operations in MAM respectively. GAP and GN are identical with that in PAM.}\label{fig_mask_att}
\end{figure}

With the introduction of contextual cues by PAM, background branch is encouraged to focus more on the regions of {\em stuff}. However, the predicted coarse areas from RPN branch lack enough cues for precise BG representations. Unlike RPN features, the $m \times m$ fixed-shape masks generated from foreground branch encode finer FG layouts. Thus, we propose Mask Attention Module (MAM) to further model the relationship, as illustrated in Figure \ref{fig_mask_att}. Consequently, the $1 \times W' \times H'$ shape FG segmentation mask is needed for similar attention operations as before. Now, the problem is: how to reproduce the $W' \times H'$ shape FG feature map from $m \times m$ masks?
%------------------------------------------------------------------------

\begin{figure}[tbp]
\centering
\subfigure[RoIAlign process]{
\includegraphics[width=0.8\linewidth]{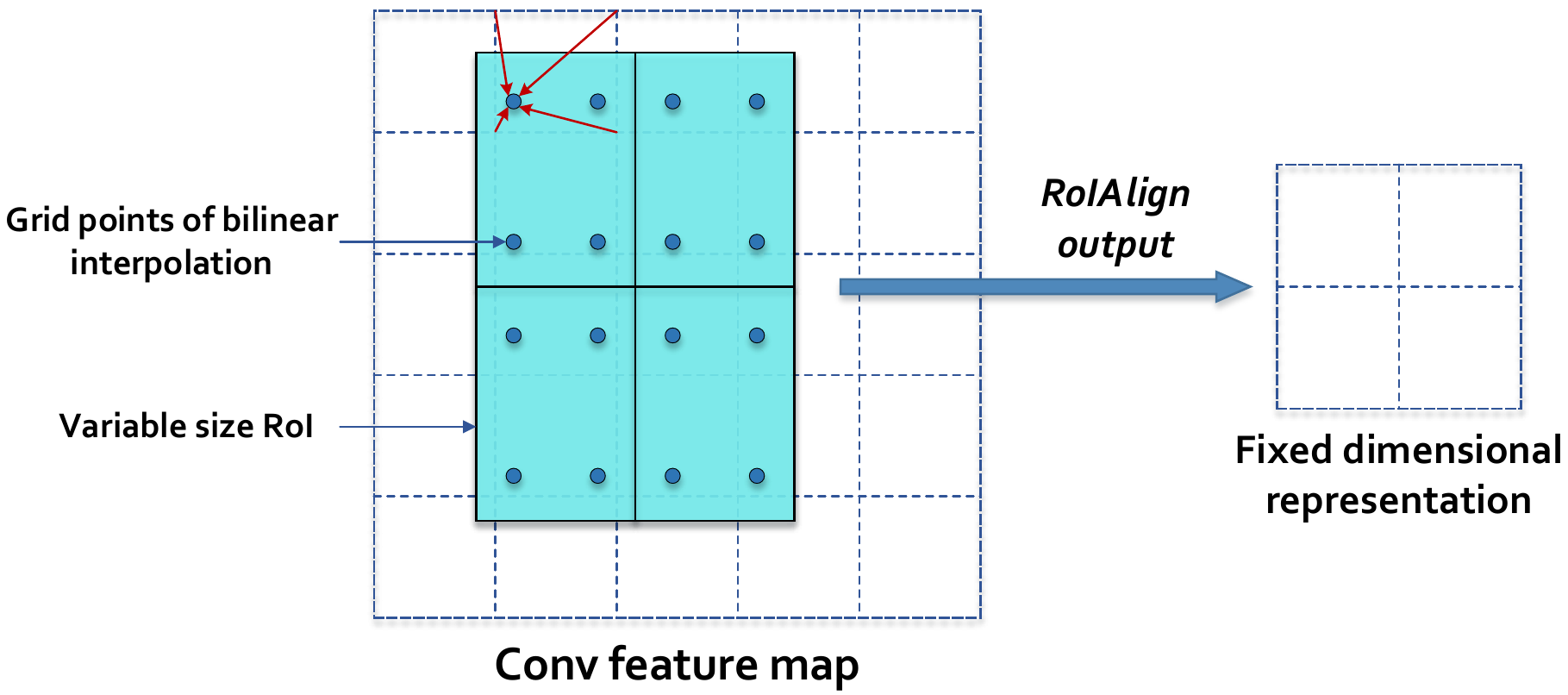}
%\caption{fig1}
}
\quad
\subfigure[RoIUpsample process]{
\includegraphics[width=0.8\linewidth]{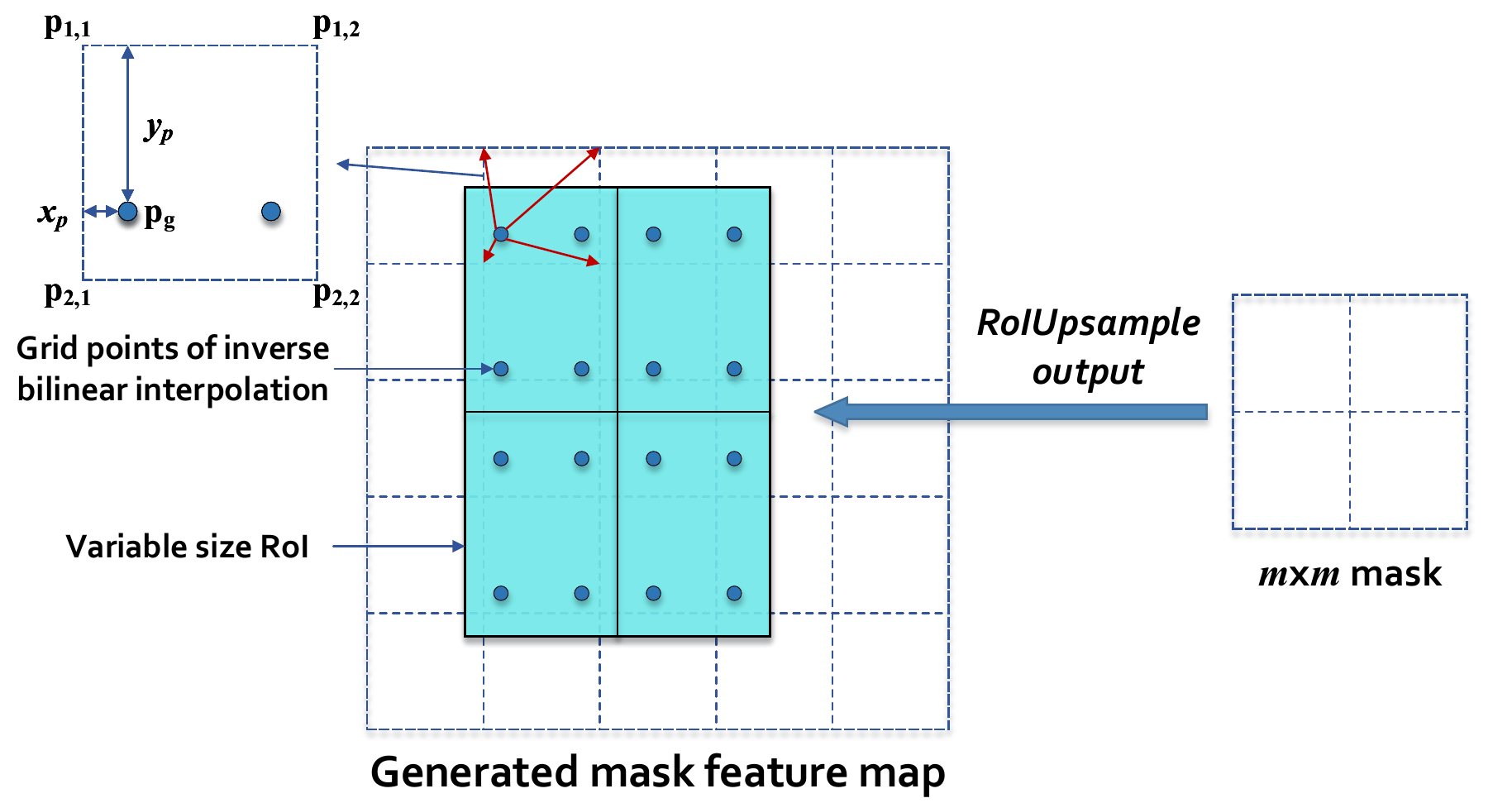}
%\caption{fig1}
\label{fig_roi_upsample}
}
\caption{Comparison between RoIAlign \cite{he2017mask} and our proposed RoIUpsample. The designed RoIUpsample, which can be viewed as an {\em inverse} operation of RoIAlign, reverts the feature map from FG masks according to their accurate spatial locations. Here, we show an example of RoIAlign output and RoIUpsample input with $m=2$ for an intuitive illustration.}
\label{fig_roi_process}
\end{figure}

\vspace{0.05cm}
\noindent{\bf RoIUpsample:} In order to solve the size mismatching problem, we propose a new differentiable layer called {\em RoIUpsample}. Specifically, RoIUpsample is designed similar to the inverse process of RoIAlign \cite{he2017mask}, as clearly illustrated in Figure \ref{fig_roi_process}. In the RoIUpsample layer, the $m \times m$ mask ($m$ equals to 14 or 28 in Mask R-CNN) is firstly reshaped to the same size of RoIs (generated from RPN). Then we utilize the designed inverse bilinear interpolation to compute values of the output features at four regularly sampled locations (same with RoIAlign) in each mask bin, and then sum up the final results as the generated mask feature map. To meet the requirement of bilinear interpolation~\cite{jaderberg2015spatial}, in which near points are given more contributions, an operation for {\em inverse} bilinear interpolation is formulated:
%------------------------------------------------------------------------
\begin{equation} \label{equ_roi_up}
\left\{ \begin{array}{lcl}
R({{\rm{p}}_{1,1}}) = \frac{{(1 - {x_p})(1 - {y_p})}}{{{{{\mathop{\rm value}\nolimits} }_x} \times {{{\mathop{\rm value}\nolimits} }_y}}}R({{\rm{p}}_g})\\
R({{\rm{p}}_{1,2}}) = \frac{{(1 - {x_p}){y_p}}}{{{{{\mathop{\rm value}\nolimits} }_x} \times {{{\mathop{\rm value}\nolimits} }_y}}}R({{\rm{p}}_g})\\
R({{\rm{p}}_{2,1}}) = \frac{{{x_p}(1 - {y_p})}}{{{{{\mathop{\rm value}\nolimits} }_x} \times {{{\mathop{\rm value}\nolimits} }_y}}}R({{\rm{p}}_g})\\
R({{\rm{p}}_{2,2}}) = \frac{{{x_p}{y_p}}}{{{{{\mathop{\rm value}\nolimits} }_x} \times {{{\mathop{\rm value}\nolimits} }_y}}}R({{\rm{p}}_g})
\end{array} \right.
\end{equation}
%------------------------------------------------------------------------
where $R({{\rm{p}}_{j,k}})$ denotes the result of point ${\rm{p}}_{j,k}$ after inverse bilinear interpolation, $R({{\rm{p}}_{g}})$ here equals to one quarter of the corresponding value in the input mask, and normalized weights ${{{\mathop{\rm value}\nolimits} }_x}$, ${{{\mathop{\rm value}\nolimits} }_y}$ are defined as:
%------------------------------------------------------------------------
\begin{equation} \label{equ_roi_up_norm}
{{\mathop{\rm value}\nolimits} _x} = x_p^2 + \left( {1 - x_p} \right)^2,{{\mathop{\rm value}\nolimits} _y} = y_p^2 + \left( {1 - y_p} \right)^2
\end{equation}
%------------------------------------------------------------------------
in which $x_p$ and $y_p$ indicate the distance between grid point ${\rm{p}}_{g}$ and generated ${\rm{p}}_{1,1}$ in two axes respectively, as presented in Figure \ref{fig_roi_upsample}. Note that with the Equation \ref{equ_roi_up} and \ref{equ_roi_up_norm}, the $m \times m$ mask can also be reverted from the generated $W' \times H'$ feature map with the {\em forward} bilinear interpolation.
%------------------------------------------------------------------------

%------------------------------------------------------------------------
Then, the generated feature map is assigned to four different scales according to the size of RoIs, which is similar with that in FPN \cite{lin2017feature}. Consequently, the generated FG feature map is achieved for the following operations.

%------------------------------------------------------------------------
\vspace{0.05cm}
\noindent{\bf Attention Operation:}  Different from traditional instance segmentation tasks, the predicted FG masks are utilized to give background branch more contextual guidance in pixel-level. We firstly aggregate them together to the $C_m \times W' \times H'$ feature map using RoIUpsample, as presented in Figure \ref{fig_mask_att}. Then, the finer $1 \times W' \times H'$ activated BG regions can be produced, similar with that in PAM. With the introduction of attention, the FG masks is also supervised by semantic loss function, which enables a further improvement in scene understanding (both for {\em things} and {\em stuff}), as discussed in Section \ref{sec::abla_study}. A similar background reweight function is adopted to aggregate useful highlighted background features.  Consequently, we model the complementary relationship between FG  {\em things} and BG {\em stuff} with the proposed PAM and MAM.
%------------------------------------------------------------------------
%-------------------------------------------------------------------------
\subsection{Implementation Details} \label{sec::implem_detail}
In this section, we give more implementation details on the training and inference stage of our proposed AUNet. 

%------------------------------------------------------------------------
\vspace{0.1cm}
\noindent{\bf Training:} As well elaborated in Section \ref{sec::net_struct}, all of our proposed methods are trained in a unified framework. The whole network is optimized via a joint loss function $\mathcal{L}$ during training stage:
\begin{equation}
\begin{split}
\mathcal{L} &= {\lambda _1}{\mathcal{L}_\mathrm{RPN}} + {\lambda _2}{\mathcal{L}_\mathrm{RCNN}} + {\lambda _3}{\mathcal{L}_\mathrm{Mask}} + {\lambda _4}{\mathcal{L}_\mathrm{Seg}}
\end{split}
\end{equation}
%------------------------------------------------------------------------
where $\mathcal{L}_\mathrm{RPN}$, $\mathcal{L}_\mathrm{RCNN}$, $\mathcal{L}_\mathrm{Mask}$ , and $\mathcal{L}_\mathrm{Seg}$ denotes the loss function of RPN, RCNN, instance segmentation, and semantic segmentation, respectively. Specifically, hyperparameters are designed to balance training processes, where $\lambda _1$ to $\lambda _4$ are set to \{1, 1, 1, 0.3\} for MS-COCO and \{1, 0.75, 1, 1\} for Cityscapes.

In details, we adopt ResNet-FPN \cite{he2016deep, lin2017feature} as our backbone. And the hyperparameters in the foreground branch are set following Mask R-CNN \cite{he2017mask}. The backbone is pre-trained on ImageNet \cite{russakovsky2015imagenet}, and the remaining parameters are initialized following \cite{he2015delving}. As standard practice \cite{he2016deep, lin2017feature, dai2017deformable}, 8 GPUs are used to train all the models. Each mini-batch has 2 images per GPU for ResNet-50 and ResNet-101 based networks and 1 image per GPU for the others. The networks are optimized for several epochs (18 for MS-COCO and 100 for Cityscapes) using mini-batch stochastic gradient descent (SGD) with a weight decay of 4e-5 and a momentum of 0.9. Batch Normalization \cite{ioffe2015batch} in the backbone is fixed and Group Normalization \cite{Wu_2018_ECCV} is added to all of the branches in our final results. For {\bf MS-COCO}~\cite{lin2014microsoft}, the learning rate is initialized with 0.02 for the first 13 epochs and divided by 10 at 15-th and 18-th epoch respectively.  Input images are horizontally flipped and reshaped to the scale with a 600 pixels short edge during training. Multi-scale testing is adopted for final results \ref{final_results}. For {\bf Cityscapes}~\cite{cordts2016cityscapes}, the learning rate is initialized with 0.01 and divided by 10 at 68-th and 88-th epoch respectively. We construct each minibatch for training from 16 random 512$\times$1024 image crops (2 crops per GPU) after randomly flipping and scaling each image by 0.5 to 2.0$\times$. Multi-scale testing is dropped in \ref{final_results}.

%-------------------------------------------------------------------------
\vspace{0.1cm}
\noindent{\bf Inference:} The panoptic results are produced in inference stage by fusing the results of FG {\em things} and BG {\em stuff} in a similar way with that in \cite{kirillov2018panoptic}. In this stage, the overlaps of {\em things} are first resolved in a NMS-like procedure which predicts the segments with higher confidence scores. And the relationships among categories are also considered during this procedure. For example, {\em ties} should not be overlapped by {\em person} in the final result. Then, the non-overlapping instance segments are combined with {\em stuff} results by assigning instance label first in favor of the {\em things}.
%------------------------------------------------------------------------
\section{Experiments}
\label{sec:experiments}
In this section, our approach is evaluated on Microsoft COCO~\cite{lin2014microsoft} and Cityscapes~\cite{cordts2016cityscapes} datasets. We first give description of the datasets as well as the evaluation metrics. Then we evaluate our method and give detailed analyses. Comparison with the state-of-the-art methods in panoptic segmentation are presented at last.
%-------------------------------------------------------------------------
\subsection{Dataset and Metrics}
\noindent{\bf Dataset:} Due to the novelty of panoptic task itself, there are few datasets with detailed panoptic annotations as well as public evaluation metrics. {\bf Microsoft COCO}~\cite{lin2014microsoft} is the most suitable and challenging one for the new panoptic segmentation task, for the detailed annotations and high data complexity. It consists of 115k images for training and 5k images for validation, as well as 20k images for {\em test-dev} and 20k images for {\em test-challenge}. MS-COCO panoptic annotations includes 80 {\em thing} categories and 53 {\em stuff} categories. We train our models on {\em train} set with no extra data and reports results on {\em val} set  and {\em test-dev} set for comparison. {\bf Cityscapes}~\cite{cordts2016cityscapes} dataset is adopted to further illustrate the effectiveness of the proposed method. In detail, it contains 2975 images for training, 500 images for validation and 1525 images for testing with {\em fine} annotations. It has another 20k {\em coarse} annotations for training, which are not used in our experiment. We report our results on {\em val} set with 19 semantic label and 8 annotated instance categories.

%-------------------------------------------------------------------------
\vspace{0.1cm}
\noindent{\bf Evaluation Metrics:} \label{eval_metric} We adopt the evaluation metrics introduced by \cite{kirillov2018panoptic}, which computes {\em panoptic quality} (PQ) metric for evaluation. PQ can be explained as the multiplication of a {\em segmentation quality} (SQ) and a {\em recognition quality} (RQ) term:

%------------------------------------------------------------------------
\begin{footnotesize}
\begin{equation}
{\rm{PQ}} = \underbrace {\frac{{\sum\nolimits_{\left( {p,g} \right) \in TP} {{\rm{IoU}}\left( {p,g} \right)} }}{{\left| {TP} \right|}}}_{{\rm{segmentation \ quality (SQ)}}} \times \underbrace {\frac{{\left| {TP} \right|}}{{\left| {TP} \right| + {\textstyle{1 \over 2}}\left| {FP} \right| + {\textstyle{1 \over 2}}\left| {FN} \right|}}}_{{\rm{recognition \ quality (RQ)}}}
\end{equation}
\end{footnotesize}
%------------------------------------------------------------------------
where ${{\rm{IoU}}( {p,g})}$ means the intersection-over-union between predicted object $p$ and ground truth $g$, true positives ($TP$) denotes matched pairs of segments (${\rm{IoU}}(p,g) > 0.5$), false positives ($FP$) represents unmatched predicted segments, and false negatives ($FN$) means unmatched ground truth segments. PQ, SQ, and RQ of both {\em thing} and {\em stuff} are also reported in our results.
%------------------------------------------------------------------------
\subsection{Component-wise Analysis and Diagnosis}
\label{sec::abla_study}
In this section, we will decompose our approach step-by-step to reveal the effect of each component. All experiments in this section are trained and evaluated on MS-COCO dataset in a single model with no extra data. Here, we adopt ResNet-50-FPN as our backbone. For fair comparison, we strictly follow the merging method in~\cite{kirillov2018panoptic} with no trick or multi-scale data augmentation in training and inference stage when doing component-wise analyses. As presented in Table~\ref{tab_ablation}, our proposed AUNet achieve an absolute improvement of $\mathbf{2.4\%}$ in PQ when compared with separate training method.

%------------------------------------------------------------------------
\begin{table}[t!]
\centering
 \caption{Comparison among different settings of panoptic quality ($\%$) on the MS-COCO dataset.  ``rewt'' means using background reweight function in PAM and MAM. PQ$^{\rm{Th}}$ and PQ$^{\rm{St}}$ indicates PQ for {\em things} and {\em stuff} respectively.}\label{tab_ablation}
 \resizebox{0.48\textwidth}{15mm}{
%\begin{footnotesize}
\begin{tabular}{lcccccccccc}
  \toprule
  Method   & PAM & MAM & rewt & PQ & PQ$^{\rm{Th}}$ & PQ$^{\rm{St}}$ & AP & mIoU \\
  \midrule
  sep   & \xmark & \xmark & \xmark & 37.2		& 47.1    & 22.8		& 33.4 & 44.5	\\
  e2e 	 & \xmark & \xmark & \xmark & 38.3	& 47.9 	& 23.9	& 33.7 & 44.8	 \\
  PAM  & \cmark & \xmark & \xmark & 39	 	& 48.5 	& 24.5	& 34.2  & 45.1	 \\
  PAM$\rm _{r}$  & \cmark & \xmark & \cmark & 39.4	& 48.9   & 25.2		& 34.4  & \textbf{45.3}	\\
  MAM  & \xmark & \cmark & \xmark & 38.9	& 48.6    & 24.2    & 34.3  & 45.2	\\
  MAM$\rm _{r}$  & \xmark & \cmark & \cmark & 39.2	& 48.6   & 24.9	 & 34.3  & 45.3	 \\
  \textbf{AUNet}  & \cmark & \cmark & \cmark & \textbf{39.6}	& \textbf{49.1}   & \textbf{25.2}  & \textbf{34.7} & 45.1	\\
  \bottomrule
 \end{tabular}
 %\end{footnotesize}
 }
\end{table}

%------------------------------------------------------------------------
\subsubsection{Unified Framework}
As elaborated in Section~\ref{sec::net_struct}, our proposed unified framework deals with FG {\em things} and BG {\em stuff} in parallel branches. As shown in Table~\ref{tab_ablation}, the unified framework boosts up the performance both in PQ$^{\rm{St}}$ and PQ$^{\rm{Th}}$, which brings $\mathbf{1.1\%}$ absolute improvements in PQ. This can be attributed to the shared backbone and joint optimization, with which the network is supervised to focus on more discriminative features for both {\em things} and {\em stuff}. With the shared backbone, the misclassification in {\em stuff} are effectively reduced and the {\em things} are given more details.

%------------------------------------------------------------------------
\subsubsection{Proposal Attention Module}
The proposed PAM builds the complementary relationship between {\em things} and {\em stuff} from different scales. By this way, the binary-classified RPN branch is optimized under the supervision of semantic labels.  With the bond between {\em stuff} and {\em things} established, the network performs consistent gain in PQ$^{\rm{St}}$ and PQ$^{\rm{Th}}$, as presented in Table \ref{tab_ablation}. The background reweight function proves its effectiveness in PQ$^{\rm{St}}$. This can be resulted from the global contextual features introduced by global average pooling in Equation~\ref{equ::pam_gap}, which means it chooses to aggregate highlighted BG features under the guidance of global context. As shown in Figure~\ref{fig_att_mask}, the activated feature map $M'_4$ emphasize the background areas with context cues. It is worth noting that we have tried other fusion methods for FG and BG feature fusion, such as concatenation and direct summary after feature transformation. But these strategies have minor contributions, which means the attention is more appropriate for relationship establishment.

\subsubsection{Mask Attention Module}
\begin{figure}[tb!]
  \centering
  \includegraphics[width=\linewidth]{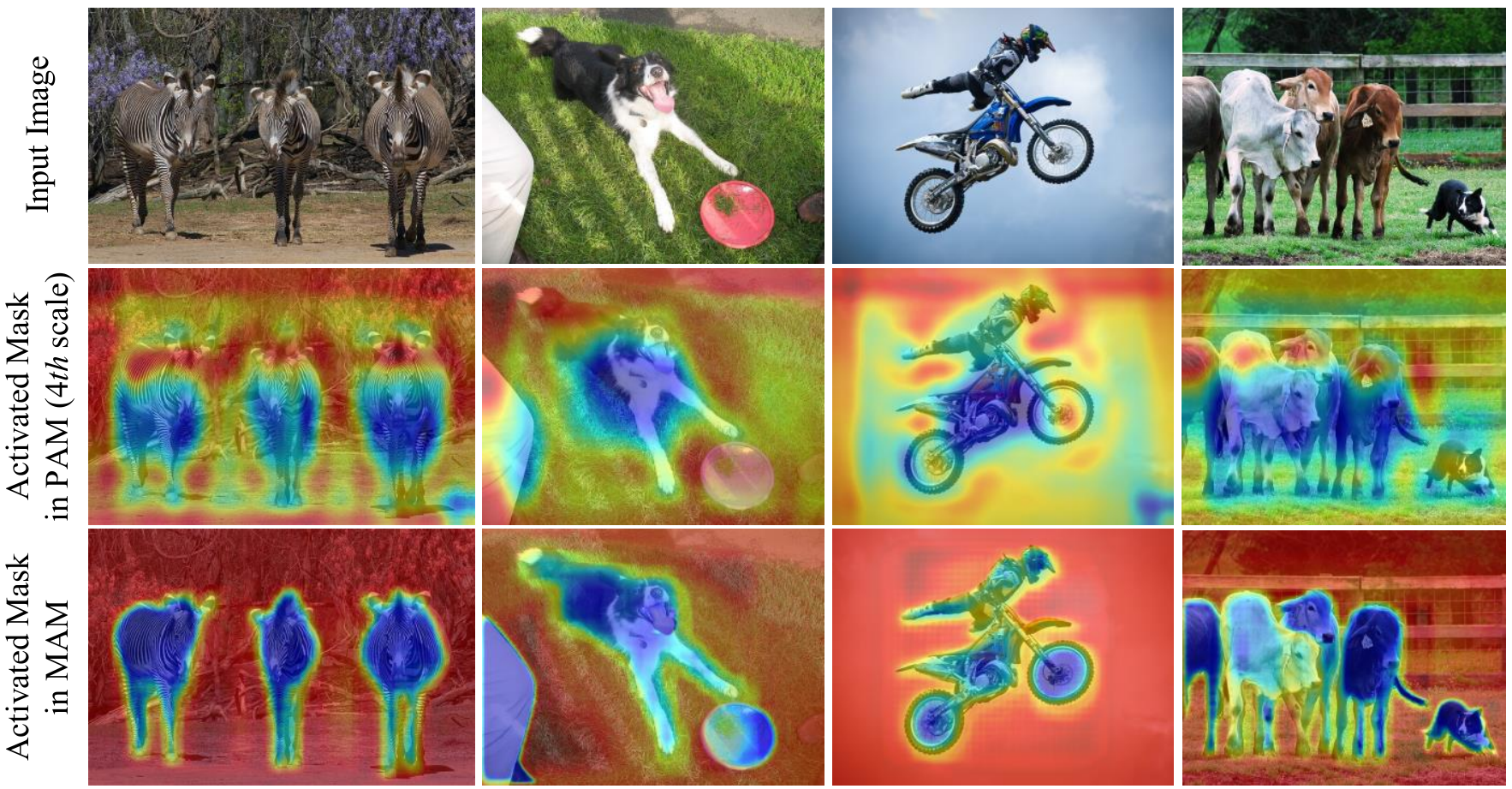}\\
  \caption{Heatmaps of the activated BG areas in PAM (the 4{\em th} scale, $M'_4$) and MAM. The red regions are assigned more weights while the blue regions less weights in the {\em background branch}. All the input images are sampled from the MS-COCO {\em val} set. }\label{fig_att_mask}
\end{figure}

While the PAM establishes the bond between FG objects and BG contents, the MAM gives background finer representations, as elaborated in Section~\ref{sec::mask_att_sec} and Figure~\ref{fig_att_mask}. As that in PAM, MAM also achieves better performance over the raw method in both PQ$^{\rm{St}}$ and PQ$^{\rm{Th}}$. However, the contribution of MAM is slightly lower than PAM. We guess this is caused by the lack of contextual cues in the generated FG segmentation mask.\footnote{We adopt zero padding for vacant areas in RoIUpsample layer, resulting in blank BG context. This needs to be investigated in the future works.} In fact, we also evaluate the performance when adopting different resolution masks for RoIUpsample, namely the $14 \times 14$ mask and the $28 \times 28$ one. The result shows the high resolution mask features bring a further gain ($0.1\%$ absolute improvement in PQ) over the smaller one. This is reasonable, because RoIUpsample layer generates finer layouts if given higher resolution masks. With the help of background reweight function, MAM$\rm _r$ achieves $39.2\%$ in PQ.
%------------------------------------------------------------------------
%----------------------------------------------------------------------

\begin{table*}[tbh]
\centering
 \caption{Panoptic quality ($\%$) on MS-COCO 2018 {\em test-dev}. ``extra data'' here denotes using extra dataset for training, ``e2e'' represents using a unified framework for {\em things} and {\em stuff} prediction, and ``enhance$_{\rm{Th}}$'' and ``enhance$_{\rm{St}}$'' indicates using additional enhancement techniques in network heads for {\em things} and {\em stuff} respectively. PQ$^{\rm{Th}}$ and PQ$^{\rm{St}}$ means PQ result for {\em things} and {\em stuff} respectively. We report our single model results with {\em  no} extra data or network enhancement.} \label{tab_main_result}
%\begin{small}
\resizebox{\textwidth}{17mm}{
\begin{tabular}{lcccccccccccccc}
  \toprule
  Method & backbone  & extra data  & e2e & enhance$_{\rm{Th}}$ & enhance$_{\rm{St}}$ & PQ & SQ & RQ &  PQ$^{\rm{Th}}$ & SQ$^{\rm{Th}}$ & RQ$^{\rm{Th}}$ & PQ$^{\rm{St}}$ & SQ$^{\rm{St}}$ & RQ$^{\rm{St}}$ \\
  \midrule
  Megvii (Face++)     & ensemble model   & \cmark & \xmark & \cmark  & \cmark & 53.2 & 83.2 & 62.9 & 62.2 & 85.5 & 72.5 & 39.5 & 79.7 & 48.5  \\
  Caribbean			  & ensemble model     & \xmark & \xmark & \cmark & \cmark   & 46.8 & 80.5 & 57.1 & 54.3 & 81.8 & 65.9 & 35.5 & 78.5 & 43.8  \\
  PKU\_360			  & ResNeXt-152-FPN	   & \xmark & \xmark & \cmark & \cmark   & 46.3 & 79.6 & 56.1 & 58.6 & 83.7 & 69.6 & 27.6 & 73.6 & 35.6 \\
  \midrule
  JSIS-Net \cite{de2018panoptic} & ResNet-50 & \xmark & \cmark & \xmark  & \xmark & 27.2 & 71.9 & 35.9 & 29.6 & 71.6 & 39.4 & 23.4 & 72.3 & 30.6 \\
 \midrule
  \textbf{Ours}   	  & ResNet-101-FPN  	   & \xmark & \cmark & \xmark  & \xmark & 45.2 & 80.6	& 54.7	& 54.4	& 83.3 	& 64.8	& 31.3 	& 76.6 	 & 39.4       \\
  \textbf{Ours}   	  & ResNet-152-FPN  	   & \xmark & \cmark & \xmark & \xmark & 45.5 & 80.8	& 55.0	& 54.7	& 83.4 	& 65.2	& 31.6 	& 76.9	 & 39.7      \\
  \textbf{Ours}   	  & ResNeXt-152-FPN    & \xmark & \cmark & \xmark & \xmark & {\bf 46.5} & {\bf 81.0}	& {\bf 56.1}	& {\bf 55.8}	& {\bf 83.7}	& {\bf 66.3}	& {\bf 32.5}	& {\bf 77.0}   & {\bf 40.7}   \\
  \bottomrule
 \end{tabular}
 }
 %\end{small}
\end{table*}
%------------------------------------------------------------------------
\begin{table}[tbh]
\centering
 \caption{Panoptic quality ($\%$) on the Cityscapes {\em val} set. PQ$^{\rm{Th}}$ and PQ$^{\rm{St}}$ denotes PQ result for {\em things} and {\em stuff} respectively. We compare our results with the bottom-up methods (the first row). Ours$_{\rm{equ}}$ indicates all {\em things} are considered as one category in the background branch during training.} \label{tab_city_result}
\resizebox{0.48\textwidth}{16mm}{
\begin{tabular}{lcccccc}
  \toprule
  Method & backbone & PQ & PQ$^{\rm{Th}}$ & PQ$^{\rm{St}}$ & AP & mIoU \\
  \midrule
  DWT~\cite{bai2017deep} & VGG16 & - & - & - & 21.2 & - \\
  SGN~\cite{liu2017sgn} & VGG16 & - & - & - & 29.2 & - \\
  Li {\em et. al.}~\cite{li2018weakly} & ResNet-101 & 53.8 & 42.5 & 62.1 & 28.6 & - \\
  \midrule
  Mask R-CNN~\cite{he2017mask} & ResNet-50 & - & - & - & 31.5 & - \\
  \midrule
  {\bf Ours$_{\rm{equ}}$} & ResNet-50-FPN & 55.0 & 51.2 & 57.8 & 32.2 & - \\
  {\bf Ours} & ResNet-50-FPN & 56.4 & 52.7 & 59.0 & 33.6 & 73.6 \\
  {\bf Ours} & ResNet-101-FPN & {\bf 59.0} & {\bf 54.8} & {\bf 62.1} & {\bf 34.4} & {\bf 75.6} \\
  \bottomrule
\end{tabular}
}
\end{table}

%------------------------------------------------------------------------
\subsection{Comparison to State-of-the-arts}\label{final_results}
We compare our proposed network with other state-of-the-art methods on MS-COCO~\cite{lin2014microsoft} {\em test-dev} and Cityscapes~\cite{cordts2016cityscapes} {\em val} set. 

\vspace{0.05cm}
\noindent{\bf MS-COCO:} As shown in Table~\ref{tab_main_result}, the proposed AUNet achieves the leading PQ performance \bm{$46.5\%$} in MS-COCO dataset without bells-and-whistles. In details, winners of COCO2018 panoptic challenge~\cite{coco2018lead} adopt numerous additional network enhancements during training and inference stage, {\em e.g.}, abundant extra data (110k external annotated MS-COCO images), multi-scale training, model ensemble. Moreover, considering the network enhancements adopted by the winner teams, cascade R-CNN~\cite{Cai_2018_CVPR} is adopted for {\em things} and extra blocks or label bank~\cite{hu2017labelbank} are added for {\em stuff} as well. Different from them, the proposed AUNet achieves the top performance in a unified framework with no extra data or additional network enhancements for both {\em things} and {\em stuff}. To be more specific, only one single model based on the ResNeXt-152-FPN\footnote{We use the $64 \times 4$d variant of ResNeXt~\cite{xie2017aggregated} with deformable conv~\cite{dai2017deformable} and non-local blocks~\cite{wang2018non}.} is adopted in the AUNet.

\begin{figure}[htbp] 
%\centering
\subfigure[Input image]{
\begin{minipage}[t]{0.3\linewidth}
\centering
\includegraphics[width=\linewidth]{Fig/main_results/raw/004.jpg}\\
\includegraphics[width=\linewidth]{Fig/main_results/raw/002.jpg}\\
\includegraphics[width=\linewidth]{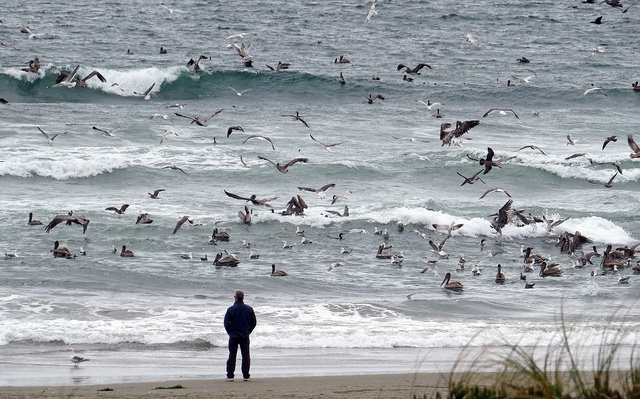}
\end{minipage}
\label{fig_final_input}
}
\subfigure[Ground truth]{
\begin{minipage}[t]{0.3\linewidth}
\centering
\includegraphics[width=\linewidth]{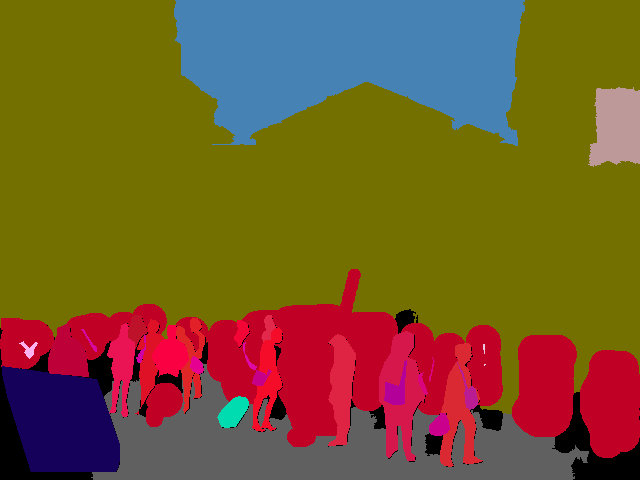}\\
\includegraphics[width=\linewidth]{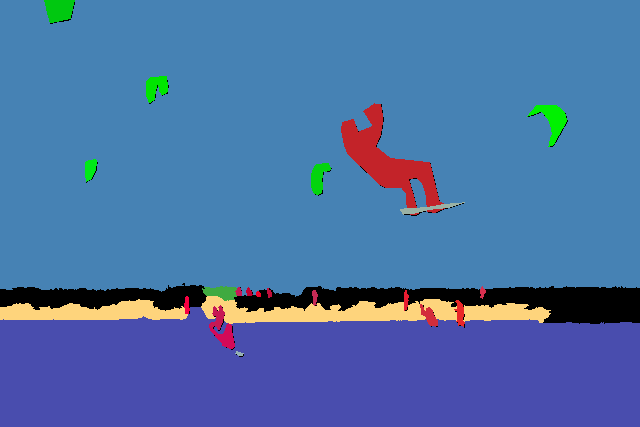}\\
\includegraphics[width=\linewidth]{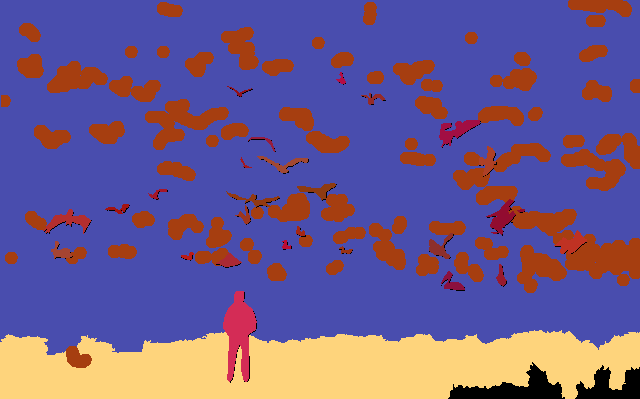}
\end{minipage}
\label{fig_final_gt}
}
\subfigure[Our results]{
\begin{minipage}[t]{0.3\linewidth}
\centering
\includegraphics[width=\linewidth]{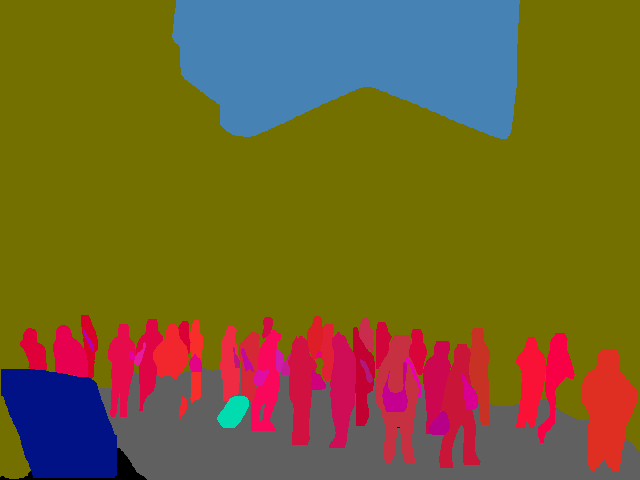}\\
\includegraphics[width=\linewidth]{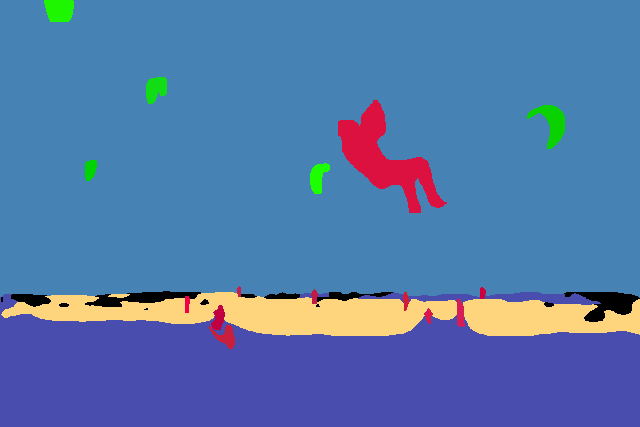}\\
\includegraphics[width=\linewidth]{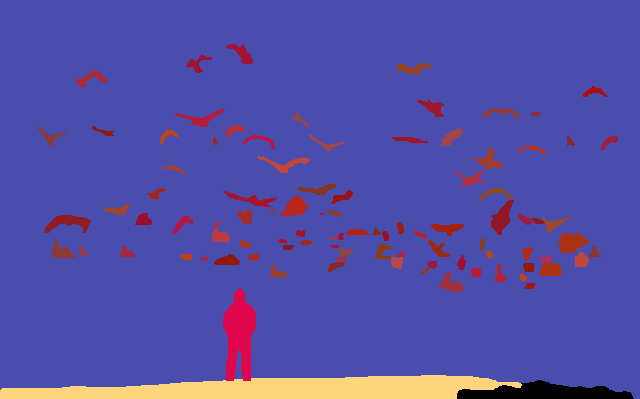}
\end{minipage}
\label{fig_final_predict}
}
\caption{Example results of AUNet on the MS-COCO {\em val} set. Our performance on {\em things}~\ref{fig_final_predict} is even better than human annotations~\ref{fig_final_gt}. The {\em things} of the same class share the same color family but appear in different intensities.}
\label{fig_final_results}
\end{figure}

Filtering out the improvement bring by model ensemble, we compare the AUNet with ``PKU\_360'' team who adopted a similar backbone but with additional skills. The result shows that our algorithm perform better than them especially in PQ$^{\rm{St}}$, for about \bm{$4.9\%$} absolute improvements. Furthermore, the AUNet overpasses the former end-to-end method, namely JSIS-Net \cite{de2018panoptic}, with a \bm{$19.3\%$} absolute gap, which proves the effectiveness of the proposed method. In Table~\ref{tab_main_result}, it is clear that the AUNet have a great balance between {\em things} and {\em stuff}, even when comparing with the challenge winners (no extra data). This is due to the introduction of unified framework and attention-guided modules for complementary relationship establishment, as well elaborated in Section~\ref{sec::abla_study}. Figure~\ref{fig_final_results} gives intuitive presentations of the top performance using our proposed AUNet.

\vspace{0.05cm}
\noindent{\bf Cityscapes:} We compare our proposed method with the leading bottom-up methods and Mask R-CNN in Table~\ref{tab_city_result}. Firstly, we adopt the same training strategy with that in MS-COCO, which means all {\em things} are considered as {\em one} category in background branch, denoted as {\bf Ours$_{\rm{equ}}$}. However, the strategy is inferior to that when using all 19 semantic labels, as illustrated in Table~\ref{tab_city_result}. Additionally, the MAM, which is proved to decrease the PQ in Cityscapes, is disabled in the final results. We guess the decline is caused by the inconsistency with prior information \ref{footnote_pri_info}, which means the relatively  worse {\em things} prediction may give wrong cues to {\em stuff}. Overall, the proposed method surpass previous state-of-the-art~\cite{li2018weakly}, with a 5.2\% absolute gap.
%------------------------------------------------------------------------
%------------------------------------------------------------------------
\section{Conclusions}
\label{sec:conclusions}
This paper presents AUNet, a unified framework for panoptic segmentation. The key difference from prior approaches lies in that we unify FG (instance-level) and BG (semantic-level) segmentation into one model, so that the FG branch, often being better optimized, can assist the BG branch via two sources of attention ({\em i.e.}, proposal attention module and mask attention module), which offer object-level and pixel-level guidance, respectively. In experiments, we observe consistent accuracy gain in MS-COCO, based on which new state-of-the-arts are achieved.

Our research delivers an important message: in visual tasks, it is often beneficial to partition targets into a few subclasses according to their properties, so that complementary information can be propagated across subclasses to assist scene understanding. Panoptic segmentation, being a new task, offers a natural partition between FG {\em things} and BG {\em stuff}, yet more possibilities remain unexplored and to be studied in the future.
%------------------------------------------------------------------------

\section*{Acknowledgement}
We would like to thank Jiagang Zhu and Yiming Hu for valuable discussions. This work was supported by the National Key Research and Development Program of China No. 2018YFD0400902 and National Natural Science Foundation of China under Grant 61573349.

\clearpage

{\small
\bibliographystyle{ieee_fullname}
\bibliography{aunet_bib}
}

\end{document}